\documentclass[journal]{IEEEtran}
\usepackage{cite}

\ifCLASSINFOpdf
   \usepackage[pdftex]{graphicx}
\else
\fi
\usepackage{amsmath}
\usepackage{algorithmic}

\usepackage{array}

  \usepackage[caption=false,font=footnotesize]{subfig}
\usepackage{url}

\usepackage{amssymb,amsmath,amsthm,amsfonts}
\usepackage{multirow}
\usepackage{enumitem}
\usepackage{hyperref}
\usepackage{color,soul}
\usepackage{bm}
\usepackage{listliketab}
\usepackage{mathtools}
\usepackage{slashbox}
\usepackage[table]{xcolor}

\newcolumntype{C}[1]{>{\centering\let\newline\\\arraybackslash\hspace{0pt}}m{#1}}
\newcolumntype{R}[1]{>{\raggedright\let\newline\\\arraybackslash\hspace{0pt}}m{#1}}
\newcolumntype{L}[1]{>{\raggedleft\let\newline\\\arraybackslash\hspace{0pt}}m{#1}}

\DeclareMathOperator*{\argmin}{arg\,min} 
\newcommand{\degree}{^\circ}
\newcommand{\etal}{et~al. }
\newcommand{\ie}{i.e. }
\newcommand{\eg}{e.g. }
\renewcommand{\eqref}[1]{(\ref{#1})}
\newcommand{\figref}[1]{Fig.\ref{#1}}

\newcommand{\subtext}[1]{_{\text{#1}}}
\newcommand{\suptext}[1]{^{\text{#1}}}

\newcommand{\inv}{^{\text{-1}}}
\newcommand{\e}[1]{E$^{\text{#1}}$}

\def \bR {\Bbb R}

\def \nn {{\mathnormal n}}

\def \nu {{\mathnormal u}}

\def \bRn {\bR^{\nn}}

\def \prox {\mathrm{prox}}
\def \diag {\mathrm{diag}}
\def \vec {\mathrm{vec}}
\def \tens {\mathrm{tens}}
\def \fold {\mathrm{fold}}

\def \vecdiag {\mathrm{vecdiag}}

\def \bR {\mathbb R}

\begin{document}
%
\title{Dynamic PET cardiac and parametric image reconstruction: a fixed-point proximity gradient approach using patch-based DCT and tensor SVD regularization}
%
%
%
\author{Ida~H\"{a}ggstr\"{o}m$^*$\thanks{$^*$I. H\"{a}ggstr\"{o}m and C. R. Schmidtlein are with the Department of Medical Physics, Memorial Sloan Kettering Cancer Center, New York, NY 10065 (e-mail: haeggsti AT mskcc DOT org).},
        Yizun~Lin\thanks{Y. Lin is with the School of Mathematics, and Guangdong Provincial Key Lab of Computational Science, Sun Yat-sen University, Guangzhou 510275, P. R. China.},
				Si~Li\thanks{S. Li is with the School of Data and Computer Science, and Guangdong Provincial Key Lab of Computational Science, Sun Yat-sen University, Guangzhou 510275, P. R. China.},
				Andrzej~Krol\thanks{A. Krol is with the Department of Radiology and Department of Pharmacology, SUNY Upstate Medical University, Syracuse NY 13210.},
				Yuesheng~Xu\thanks{Y. Xu is the the Department of Mathematics and Statistics, Old Dominion University, Norfolk, VA 23529.},
        and~C.~Ross~Schmidtlein}

\maketitle

\begin{abstract}
Our aim was to enhance visual quality and quantitative accuracy of dynamic positron emission tomography (PET) uptake images by improved image reconstruction, using sophisticated sparse penalty models that incorporate both 2D spatial+1D temporal (3DT) information.
We developed two new 3DT PET reconstruction algorithms, incorporating different temporal and spatial penalties based on discrete cosine transform (DCT) w/ patches, and tensor nuclear norm (TNN) w/ patches, and compared to frame-by-frame methods; conventional 2D ordered subsets expectation maximization (OSEM) w/ post-filtering and 2D-DCT and 2D-TNN. A 3DT brain phantom with kinetic uptake (2-tissue model), and a moving 3DT cardiac/lung phantom was simulated and reconstructed. For the cardiac/lung phantom, an additional cardiac gated 2D-OSEM set was reconstructed. The structural similarity index (SSIM) and relative root mean squared error (rRMSE) relative ground truth was investigated. The image derived left ventricular (LV) volume for the cardiac/lung images was found by region growing. Parametric images of the brain phantom were calculated by nonlinear least squares fitting.
For the cardiac/lung phantom, 3DT-TNN yielded optimal images with an rRMSE / SSIM of 45.4$\pm$0.4\% / 0.652$\pm$0.007, compared to 65.4$\pm$0.1\% / 0.4439$\pm$8\e{-4} for conventional 2D-OSEM. The optimal LV volume from the 3DT-TNN images was on average 79$\pm$4\% of the true value, cardiac gated 2D-OSEM 68$\pm$9\%, and 2D-OSEM 24$\pm$7\%. 
3DT-DCT had minimum rRMSE and maximum SSIM for the brain phantom at 59.5$\pm$0.3 and 0.593$\pm$0.003\% respectively, whereas 2D-OSEM had an rRMSE / SSIM of 75.6$\pm$0.4\% / 0.478$\pm$0.005.
 Compared to 2D-OSEM, parametric images based on 3DT-DCT images had smaller bias for all six parameters, and higher SSIM for all but one.

Our novel methods that incorporate both 2D spatial and 1D temporal penalties produced dynamic PET images of higher quality than conventional 2D methods, w/o need for post-filtering. Breathing and cardiac motion were simultaneously captured w/o need for respiratory or cardiac gating. LV volumes were better recovered, and subsequently fitted parametric images were generally less biased and of higher quality.
\end{abstract}


%
\IEEEpeerreviewmaketitle


\section{Introduction}
\noindent \IEEEPARstart{D}{ynamic} positron emission tomography (PET) is used to quantify physiological and functional processes \emph{in vivo}. In particular, these images have been used to identify perfusion or blood volume in cardiac studies, improve tumor uptake estimates in the presence of respiratory motion, and to estimate parameters used in kinetic modeling studies. In each of these cases, the dynamic nature of the data is generally acquired in static bins. In kinetic modeling these bins are predefined to try and capture the signal in a way that is related to the rate of change of activity distribution. In the cases of cardiac and respiratory motion compensation the static frames are defined through the gating information to visualize distinct phases of the cardiac or breathing cycle.
\newline\indent
In cardiac PET, the accurate measurement of the left ventricular (LV) function, volume, and mass holds important diagnostic information\cite{White1987,Lorell2000} that can help identify myocardial ischemia; useful in evaluating damage from heart attacks, surgical risk, etc. Typically, both kinetic modeling of the dynamic data along with electrocardiogram-gated data are required for full analysis of molecular processes as well as functional LV assessment. 
This generally requires separate reconstructions, significantly increasing workload, and limiting the use of combined functional and molecular measurements in cardiac imaging\cite{Harms2016}. In both cases, gated or dynamic, the data is generally acquired as static bins/frames that average the data independently over each small bin. 
Measurements in the thoracic region are typically gated, due to organ or tissue motion. Gating however, whether it be cardiac or respiratory, leads to lower count statistics for each gated image frame, and hence noisier images since correlation between the bins/frames is not considered during the image estimation process. Furthermore, very low count statistics yield biased images for conventional maximum likelihood expectation maximization (MLEM) reconstruction methods\cite{Walker2009}. Recently, Harms \etal\cite{Harms2016} were able to determine the LV mass and volumes, together with parametric images from ungated dynamic $\suptext{11}$C-acetate PET data. Long acquisition times (27~min) were required for parameter estimation using a one-tissue compartment model (parameters $K_1$ and $k_2$) for the kinetic analysis.
\newline\indent
Although useful in cardiac imaging, parametric imaging has proven especially valuable in oncology for aiding in staging, treatment planning, delineation, follow-up, and response evaluation of cancerous tumors\cite{Weber2006,Muzi2012,Clarke2009,Cheebsumon2011,Haggstrom2010}. However, the dynamic PET images, on which the parametric fits are based, generally have poor quality due to low count statistics and because inter-frame correlations are ignored. This occurs because the time-dependent aspects of the tracer distribution are independent of the ML reconstruction model. As a result, dynamic PET images are independently reconstructed frame-by-frame, where time correlation, both intra- and inter-frame, are ignored leading to spatially and temporally noisy dynamic images.
\newline\indent
In each case described above noise degrades the utility of the data. The finer the desired resolution, be it spatial, temporal, or parametric (more model detail), the more noise limits our ability to accurately estimate the activity distribution (parametrized or otherwise). To improve estimation, a number of denoising approaches have been investigated. Broadly speaking, these fall into two general classes: denoising post-reconstruction, and denoising integrated into the reconstruction problem. 
\newline\indent
Extensive work has been done on both cases, but our interest is in the integrated reconstruction problem due to the more direct connection it formalizes between image likelihood and the denoising process. Amongst the most common are wavelet, discrete cosine transform (DCT), singular value decomposition (SVD), and other transform methods, which seek a sparse transform space where noise can be removed via thresholding \cite{Hou2003,Hongbo2016,He2011,Zhang2015}. These have been very effective with 2D images and can be readily adapted to higher dimensional images. 
\newline\indent
One interesting approach is based on tensor SVD (tSVD), a higher dimensional generalization of SVD\cite{Kilmer2013,Zhang2014}. Li \etal~\cite{Li2017} developed an iterative model-based dynamic CT reconstruction algorithm based on SVD, where they used the nuclear norm ($\ell_1$-norm of singular values) as a regularizer. Their work used a weighted least squares minimizer.
While these sparse representation methods rely on a thresholding procedure that complicates the optimization process due to its nondifferentiability, a number of algorithms for PET have been developed. For 4D PET reconstruction, methods based on \eg E-spline wavelets\cite{Verhaeghe2008}, complex wavelets\cite{McLennan2009}, total variation (TV)\cite{Chen2015}, and robust principal component analysis\cite{Yu2015} exist. 
Sparse methods connected to kinetic modeling and direct parametric reconstruction are summarized in the review articles\cite{Rahmim2009,Wang2013,Reader2014}. 
\newline\indent
This work focuses on time resolved PET reconstruction, rather than direct parametric imaging, for several reasons. While direct parametric reconstruction has been very successful in reducing the variance of derived parameters\cite{Rahmim2009,Wang2013,Kamasak2005,Reader2014}, these models are typically coupled to features found in tracer kinetics, not motion, and may perform poorly for fast dynamics or quick motion. Furthermore, the direct parametric reconstruction of model parameters is inherently tied to the choice of kinetic model. Any real deviations, such as motion or an inappropriate model choice, between the patient's data and the model will bias the reconstruction process - particularly problematic for newer tracers where uptake mechanisms are not well established. Another difficulty is that kinetic modeling typically requires an \emph{a priori} blood input function (IF), though both population-based IFs, and joint estimate of the IF and parametric images have been studied\cite{Zanotti2011,Wang2013}. In fully 4D PET reconstruction our transform space can be more general, making motion visible in the 4D images, and fitting deviation, as evidenced by a large residual, readily apparent.
We note that there are many methods for direct parametric image reconstruction, but for the reasons above these will not be addressed here.
\newline\indent
The aim of this study is to improve dynamic PET image quality and quantitative accuracy, by novel reconstruction algorithms containing improved regularization terms based on sparse representation that incorporate both spatial and temporal correlations. We believe the main benefits are 1) improved handling of regular plus irregular motion, 2) more accurate myocardial measurements, 3) improved parametric image quality and quantitative accuracy, and 4) simplified kinetic model selection for parametric imaging and improved evaluation of model suitability. We simulated multiple noise realizations of two examples - one brain phantom with dynamic uptake, and one cardiac/lung phantom with motion - and investigated image quality and succeeding calculated metrics for both conventional and novel reconstruction methods. 

\section{Methodology}
\subsection{Emission computed tomography regularization models}
Emission computed tomography (ECT) image reconstruction is based on a Poisson noise model given by
\begin{equation}
g = \text{Poisson}\{Af+\gamma\},
\label{eqn:PET}
\end{equation}
where $g\in\mathbb{Z}\subtext{+}^{N}$ is the projection data, $A\in\mathbb{R}^{N \times M}$ is the linear projection operator, $f\in\mathbb{R}\subtext{+}^{M}$ is the activity distribution, and $\gamma \in\mathbb{R}\subtext{+}^{N}$ is the additive counts from random and scatter events. The log-likelihood of this model leads to Kullback Leibler-divergence fidelity term given by $F(f)$ to which a penalty function $\Psi(f)$ with a penalty weight $\lambda$ can be attached, forming the objective function $\Phi(f)$. These are given by
\begin{align}
F(f) & \triangleq \langle A f,\textbf{1}\rangle - \langle \log{(A f + \gamma)}, g \rangle,\\
\Psi(f) & \triangleq \varphi(Bf) ,~\text{and} \\
\Phi(f) & \triangleq F(f) + \lambda \Psi(f),
\label{eqn:model}
\end{align}
where \textbf{1} is the vector of all ones, $\langle \cdot , \cdot \rangle$ denotes the inner product, and $\varphi(Bf)$ is a regularization term (typically some norm). The corresponding minimization problem is 
\begin{equation}
\begin{split}
& \min{\Phi(f)}\\
& \text{subject to}~ f \geq 0.
\end{split}
\label{eqn:PLObj}
\end{equation}
The gradient or subgradient with respect to $f$ of these functions are given by
\begin{align}
\nabla_f F & = A^{\top}\left( \textbf{1} - \frac{g}{Af + \gamma} \right),~\text{and}\\
\partial_f \Psi & = B^{\top}\partial \varphi (Bf),
\label{eqn:gradFidelity}
\end{align}
respectively, where $\varphi (\cdot)$ represents the particular norm used, and $\partial (\cdot)$ is the subdifferential.
\newline\indent
The fidelity term in physical model described by \eqref{eqn:model} is time independent.  Any additional constraints in the time domain must be applied by the regularization.  In this work, we describe fixed-point (FP) algorithms for several regularization models based on the form given by \eqref{eqn:PLObj} that include the time domain. These include DCT, and the nuclear norm (also known as the trace- and Schatten-norm) for higher-dimensional ECT. In addition, we describe patch-wise extensions of these methods for increased regularization redundancy.
\subsubsection{Selected properties of the proximity operator}
Prior to deriving the optimization algorithms for solving \eqref{eqn:PLObj} with the various regularization models the proximity operator and several of its properties should be discussed.
\newline\indent
The proximity operator of function $\psi$ is given by
\begin{equation}
\prox_{\psi}(x)\triangleq\argmin_{u\in\bRn}\left\{\frac{1}{2}\Vert u-x\Vert_2^2+\psi(u)\right\}~.
\label{eqn:prox_psi}
\end{equation}
This function can be explicitly evaluated for a number of functions and norms. In this paper we are particularly interested in the indicator function, the $l_1$-norm, and the nuclear norm.
\newline\indent
Denoting the indicator function of a convex set, $\iota\subtext{+}(f)$ (where + represents inclusion in $\mathbb{R}\subtext{+}$), its proximity can be evaluated as
\begin{equation}
P\subtext{+}(f) \triangleq \prox_{\iota\subtext{+}}(f) = \max\{f,0\},
\label{eqn:prox_indicator}
\end{equation}
where $\max\{f,0\}\in\bR\subtext{+}^M$ and its $i$th element is given by $\max\{f_i,0\}$.
Denoting
\begin{equation}
p=\prox_{\frac{1}{\mu} \|\cdot\|_1}(f),
\end{equation}
then the explicit form for the proximity of the $\ell_1$-norm can be given by
\begin{equation}
p_i = \max \{|f_i|-\tfrac{1}{\mu},0\} \text{sign}(f_i).
\label{eqn:l1_prox}
\end{equation}
\newline\indent
In the case of dynamic (3DT) PET image $\in\bR\subtext{+}^{m\times n\times\tau}$ we expect the 3DT-images to be low rank due to low counts (possibly under determined) and because variation in the time domain will be smooth with considerable redundancy.  The nuclear norm, \ie the $l_1$-norm of the singular values denoted $\|\cdot\|_*$, is the tightest convex approximation of the rank of a 2D matrix.  However, SVD in high dimensions does not have a unique representation, and some representations are more useful than others for image processing. 
\newline\indent
Tensor SVD in particular readily generalizes to higher dimensions and has been shown to be very effective in image processing\cite{Kilmer2013,Zhang2014}. Following \cite{Kilmer2013}, the tSVD of a 3DT tensor $\mathcal{T}\in\bR^{m\times n\times \tau}$, we first use the Fourier transform on the third dimension (denoted by $\mathcal{F}_{(3)}$) of $\mathcal{T}$ to get $\hat{\mathcal{T}}$, that is, $\hat{\mathcal{T}}\triangleq\mathcal{F}_{(3)}(\mathcal{T})$. For each $i$, $\hat{\mathcal{T}}^{(i)}=\hat{\mathcal{T}}(:,:,i)$ is an $m\times n$ matrix which has the conventional SVD $\hat{\mathcal{T}}^{(i)}=\hat{\mathcal{U}}^{(i)}\hat{\mathcal{S}}^{(i)}\hat{\mathcal{V}}^{(i)\top}$. 
The tSVD of $\mathcal{T}$ for higher dimensional tensors was given by Zhang \etal \cite{Kilmer2013,Zhang2014}.  
\newline\indent
We now define the tensor nuclear norm (TNN) of $\mathcal{T}$ by
\begin{equation}
\|\mathcal{T}\|\subtext{TNN} \triangleq \sum^{\tau}_{i=1}{\big\lVert \vecdiag\big(\hat{\mathcal{S}}^{(i)}\big)\big\rVert_1},
\label{eqn:tNN}
\end{equation}
where for $\Lambda\in\bR^{m\times n}$, $\vecdiag(\Lambda)$ represents the vectorization of the diagonal entries of $\Lambda$, that is $\vecdiag(\Lambda)\in\bR^{\text{min}(m,n)}$ and its $j$th element is given by $\left(\vecdiag(\Lambda)\right)_j=\Lambda(j,j)$ for $j=1,2,\dots,\text{min}(m,n)$.
\newline\indent
Now for a 3DT tensor $\mathcal{W}\in\bR^{m\times n\times\tau}$, we define the vectorization of $\mathcal{W}$ as $\vec(\mathcal{W})\in\bR^{mn\tau}$, such that the $\left((k-1)mn+(j-1)m+i\right)$th element of $\vec(\mathcal{W})$=$\mathcal{W}(i,j,k)$ for $i=1,2,\dots,m$, $j=1,2,\dots,n$, and $k=1,2,\dots,\tau$. Conversely, for a vector $f\in\bR^{mn\tau}$, $\tens(f)\in\bR^{m\times n\times\tau}$ is defined as the tensor form of $f$, which is the inverse of the operator $\vec(\cdot)$. In addition, we define the TNN norm of the vector $f$ as 
\begin{equation}
\|f\|\subtext{TNN}\triangleq\|\tens(f)\|\subtext{TNN}.
\label{eqn:tNNtens}
\end{equation}
Using the above definitions, the proximity operator of the TNN at $f\in\bR^{mn\tau}$ can be given by 
\begin{equation}
\begin{aligned}
&\hskip0pt\prox_{\frac{1}{\mu} \|\cdot\|\subtext{TNN}}(f) = \argmin_{u\in\bR^{mn\tau}}\bigg\{\frac{1}{2}\|u-f\|_2^2+\frac{1}{\mu}\|u\|\subtext{TNN}\bigg\} \\
&\hskip0pt= \argmin_{u\in\bR^{mn\tau}}\bigg\{\frac{1}{2}\|\mathcal{F}_{(3)}(\tens(u))-\mathcal{F}_{(3)}(\tens(f))\|_F^2+\frac{1}{\mu}\|u\|\subtext{TNN}\bigg\} \\
&\hskip0pt= \vec\bigg(\mathcal{F}_{(3)}\suptext{-1}\argmin_{\mathcal{W}\in\bR^{m\times n\times\tau}}\Big\{\frac{1}{2}\|\mathcal{W}-\tens(f)\|_F^2+\frac{1}{\mu}\sum^{\tau
}_{i=1}{\|\mathcal{W}^{(i)}\|_{*}}\Big\}\bigg) \\
&\hskip0pt= \vec\bigg(\mathcal{F}_{(3)}\suptext{-1}\Big(\fold\Big(\big\{\prox_{\frac{1}{\mu}\|\cdot\|_{*}}\big(\tens(f)^{(i)}\big)\big\}_{i=1}^n\Big)\Big)\bigg)
\label{eqn:tNNprox}
\end{aligned}
\end{equation}
where $\|\cdot\|_F$ is the Frobenius norm defined by
\begin{equation}
\|\mathcal{T}\|_F\triangleq\Bigg(\sum_{(i,j,k)=(1,1,1)}^{(m,n,\tau)}{\mathcal{T}(i,j,k)^2}\Bigg)^{1/2},
\end{equation}
for $\mathcal{T}\in\bR^{m\times n\times\tau}$, and the operator $\fold(\cdot)$ turns $\tau$ $m\!\!\times\!\!n$ matrices into a 3D tensor as $\fold\big(\mathcal{T}^{(1)},\mathcal{T}^{(2)},\dots,\mathcal{T}^{(\tau)}\big)=\mathcal{T}$.
\newline\indent
The algorithms developed in this paper are based upon the FP proximity gradient (PG) algorithm described by \cite{Li2015b,Wu2016}. The basis of these FP equations is the subdifferential property of the proximity operator defined in \eqref{eqn:prox_psi} as described in \cite{Micchelli2011}. Qualitatively, that is that $y \in \partial \psi(x)$, where $x,y \in \bR^M$ and $\psi$ is a real valued convex function, if and only if $x = \prox_{\psi}(x+y)$. This allows the FP formulation of the subdifferential as
\begin{equation}
y = \left(I-\prox_{\psi}\right)(x+y).
\label{eqn:fixedPointProx}
\end{equation}
For the formal definition and proof of the above statements see \cite{Micchelli2011}, proposition 2.6.
Following the methodology of \cite{Li2015b,Wu2016}, the various properties of the proximity operator shown above are used in the subsections below to generate algorithms for the regularization models used in this paper.
\subsection{3DT-DCT regularization}
Using the 3DT-DCT for regularization the model becomes
\begin{equation}
\Psi(f) \triangleq \|D f\|_1,
\label{eqn:DCTpenalty}
\end{equation}
where $D$ is the 3DT-DCT transform matrix. Because the 3DT image is vectorized the formulation for higher-dimensions is implicit in the definition of $D$. Using this regularization in the model described by \eqref{eqn:PLObj} a FPPG algorithm can be given by
\begin{equation}
\begin{dcases}
f^{(k+1)} \!\!\!\!&= P\subtext{+}\Big(f^{(k)}-\beta S^{(k)}\big(\nabla_f F(f^{(k)})+\lambda D^{\top}c^{(k)}\big)\Big)\\
h^{(k+1)} \!\!\!\!&= 2f^{(k+1)}-f^{(k)}\\
c^{(k+1)} \!\!\!\!&= \mu\left(I-\prox_{\frac{1}{\mu}\|\cdot\|_1}\right)\left(\frac{1}{\mu}c^{(k)}+Dh^{(k+1)}\right)
\end{dcases},
\label{eqn:DCT-PGA}
\end{equation}
where the $P\subtext{+}$ function operates element-wise. In this study $\beta$ is a step size that we set to 1, however, when using incremental subgradient methods\cite{Nedic2001} (\eg ordered subsets) it can be utilized as a relaxation term\cite{Schmidtlein2017}. The parameter $\mu$ is given by
\begin{equation}
\mu = \frac{1}{2\lambda\|D\|_2^2 \|S^{(k)}\|_2}.
\label{eqn:mu}
\end{equation}
Due to the difference in total number of counts (noise) in each projection frame of the dynamic sequence, the weights should be scaled to account for this when reconstructing. Using a simplified approach derived from Schmidtlein \etal\cite{Schmidtlein2017} using Poisson statistics weighting, the scaling is done as 
\begin{equation}
\lambda_i = \lambda\subtext{ref}\sqrt{\bar{c}/c_i},
\label{eqn:lambda} 
\end{equation}
where $\lambda\subtext{ref}$ is a scalar reference weight, $c_i$ the total counts of frame $i$, and $\bar{c}$ the average of all frames. Thus, larger penalty for lower counts. Note that $\lambda\subtext{ref}$ is still used for calculation of $\mu$ in \eqref{eqn:mu}.
\newline\indent
The preconditioner, $S^{(k)}$, taken from the ML-EM algorithm, is a diagonal matrix with entries
\begin{equation}
S_{jj}^{(k)} = \begin{dcases}
{\max\{f_{j}^{(k)},\varepsilon\}}/{(A^T \textbf{1})_{j}}, & (A^T \textbf{1})_{j} > 0\\
\max \{f_{j}^{(k)},\varepsilon\}, & (A^T \textbf{1})_{j} \le 0
\end{dcases},
\end{equation}
and $\varepsilon$ is a small positive constant, preventing a zero-valued $S$.
\subsection{3DT-TNN regularization}
In image processing, tSVD (SVD for 2D) is very useful because is uses the data itself to generate the basis, where large singular values indicate dominant image features. These aspects of the penalty are usually expressed as a nuclear norm regularization model given by
\begin{equation}
\Psi(f) \triangleq \|f\|\subtext{TNN},
\label{eqn:NNpenalty}
\end{equation}
for $\|f\|\subtext{TNN},$ in \eqref{eqn:tNN}. This can be inserted into the minimization model given by \eqref{eqn:PLObj}, and we can describe a FP algorithm for this problem.
\newline\indent
Using a similar set of arguments for deriving \eqref{eqn:DCT-PGA} a FP algorithm for TNN regularization can be found, though some differences exist. We can write the algorithm for solving model (\ref{eqn:PLObj}), using \eqref{eqn:NNpenalty} and the relations \eqref{eqn:tNN} and \eqref{eqn:tNNprox}, as
\begin{equation}
\hspace{-0.9em}
\begin{dcases}
f^{(k+1)} &\!\!\!= P\subtext{+}\Big(f^{(k)}-\beta S^{(k)}\big(\nabla_f F(f^{(k)})+\lambda c^{(k)}\big)\Big)\\
h &\!\!\!= \tfrac{1}{\mu}c^{(k)}\left(2f^{(k+1)} - f^{(k)}\right)\\
\hat{\mathcal{H}} &\!\!\!= \mathcal{F}_{(3)}\big(\tens(h)\big)\\
\omit{\rlap{$
\left\{\hat{\mathcal{U}}^{(i)},\hat{\mathcal{S}}^{(i)},\hat{\mathcal{V}}^{(i)\top}\right\} = \text{SVD}\left(\mathcal{H}^{(i)}\right),~\forall{i}
$}}&\\
\omit{\rlap{$
\mathcal{P}^{(i)}=\hat{\mathcal{U}}^{(i)}\diag_{m\times n}\left\{ \prox_{\frac{1}{\mu}}\Big(\vecdiag\big(\hat{\mathcal
{S}}^{(i)}\big)\Big)\right\}\hat{\mathcal{V}}^{(i)\top},~\forall{i}
$}}&\\
s &\!\!\!= \vec\bigg(\mathcal{F}_{(3)}\inv\Big(\fold\big(\big\{\mathcal{P}^{(i)}\big\}_{i=1}^\tau\big)\Big)\bigg),~\forall{i}\\
c^{(k+1)} &\!\!\!= \mu\left(h-s\right)
\end{dcases}.
\label{eqn:tSVD-PGA}
\end{equation}
We note that for the 2D case, the above expression holds without the need for the higher dimension Fourier transform.

\subsection{Nonlinear filter extension}
\begin{figure}[bt!]
	\centering
		\includegraphics[width=1\columnwidth]{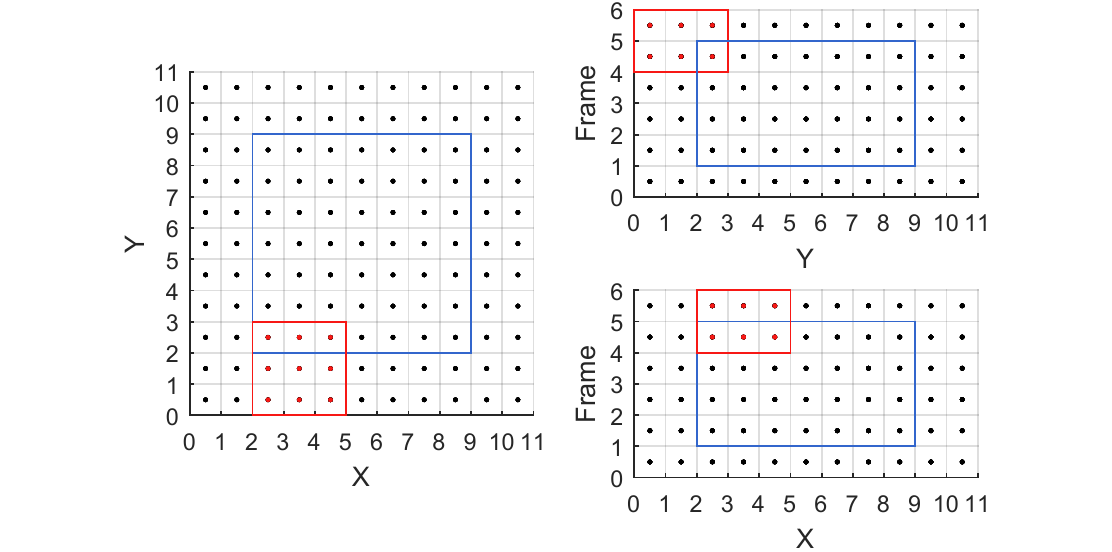}
	\caption{Example of a $7\times7\times4$ dynamic image (blue) with padding, and a $3\times3\times2$ patch (red). Dots are voxel centers.}
	\label{fig:patch}
\end{figure}

For both 3DT-DCT and TNN it is natural to consider these regularizers patch-wise as nonlinear filters that can be applied as either sliding windows or as overlapping patches to add redundancy and additional degrees of freedom to the sparse transform domain. In either case, the mathematical formalism for the patches is the same, where we begin with some patch extraction operator given by
\begin{equation}
q = Q f~\in \mathbb{R}\subtext{+}^{L}~.
\label{eqn:Qpatches}
\end{equation}
In this case, $Q \in \mathbb{R}\subtext{+}^{L \times M}$, where $L$ is the total number of elements from all patches (with $L/M$ patches), is a linear operator that redundantly samples $f$ to produce a patch vector $q$, where further operations can be applied to $q$ such as DCT or TNN. The $j$th patch is denoted $q_j=Q_jf$, and the extraction is inverted as $f=Q^{\top}q$.
The edges of images represent a special challenge were we symmetrically pad the original images by half the patch size in each direction prior to patch extraction. An example of a patch in a 3DT image is seen in \figref{fig:patch}.

\subsubsection{Patch-based 3DT-DCT regularization}\label{sec:nd-dct}
In the case of 3DT-DCT, the patches contain 3DT regions of the image. Extending the 3DT DCT transform $D$ to $D_q$ which handles patches, the regularization model is given by
\begin{align}
\Psi(f) & \triangleq \|D_q Q f\|_1,~\text{and}\\
D_q & \triangleq I_{L/M} \otimes D,
\label{eqn:qDCTpenalty}
\end{align}
where $\otimes$ is the Kronecker product, which combined with the model \eqref{eqn:PLObj} the algorithm can be directly created using \eqref{eqn:DCT-PGA}
\begin{equation}
\begin{dcases}
f^{(k+1)} &\!\!\!= P\subtext{+}\Big(f^{(k)}-\beta S^{(k)}\big(\nabla_f F(f^{(k)})\\
&\!\!\!\quad +\lambda Q^{\top}D_q^{\top} c^{(k)}\big)\Big)\\
h^{(k+1)} &\!\!\!= 2f^{(k+1)} - f^{(k)}\\
c^{(k+1)} &\!\!\!= \mu \left(I - \prox_{\frac{1}{\mu}\|\cdot\|_1}\right)\\
&\!\!\!\quad \times\Big(\tfrac{1}{\mu}c^{(k+1)} + D_q Q h^{(k+1)}\Big)
\end{dcases}.
\label{eqn:qDCT-PGA}
\end{equation}
The patch extractor $Q$ can be extended to $Q_{\circ}=QM_{\circ}$ which includes a rotation operation $M_{\circ}$ prior to extraction to further increases the redundancy. In this work, a $45\degree$ rotation in the $xy$-plane was included, adding an extra penalty term $\lambda Q_{45}^{\top}D_q^{\top} c_{45}^{(k)}$ to $f^{(k+1)}$ in \eqref{eqn:qDCT-PGA}.
 
\subsubsection{Patch-based 3DT-TNN regularization}\label{sec:nd-tnn}
In the case of 3DT-TNN, the patches contain 3DT regions of the image where the regularization model is given by
\begin{equation}
\hspace{-1em}
\begin{dcases}
f^{(k+1)} &\!\!\!= P\subtext{+}\Big(f^{(k)}-\beta S^{(k)}\big(\nabla_f F(f^{(k)})+\lambda Q^{\top}c^{(k)}\big)\Big)\\
h_j &\!\!\!= \tfrac{1}{\mu}c_j^{(k)}Q_j\left(2f^{(k+1)} - f^{(k)}\right),~\forall{j}\\
\hat{\mathcal{H}}_j &\!\!\!= \mathcal{F}_{(3)}\big(\tens(h_j)\big),~\forall{j}\\
\omit{\rlap{$
\left\{\hat{\mathcal{U}}^{(i)},\hat{\mathcal{S}}^{(i)},\hat{\mathcal{V}}^{(i)\top}\right\}_j =\text{SVD}\left(\mathcal{H}^{(i)}_j\right),~\forall{ij}
$}}\\
\omit{\rlap{$
\mathcal{P}^{(i)}_j=\hat{\mathcal{U}}_j^{(i)}\diag_{m\times n}\left\{ \prox_{\frac{1}{\mu}}\Big(\vecdiag\big(\hat{\mathcal{S}}_j^{(i)}\big)\Big)\right\}\hat{\mathcal{V}}_j^{(i)\top},~\forall{ij}
$}}\\
s_j &\!\!\!= \vec\left(\mathcal{F}_{(3)}\inv\Big(\fold\big(\big\{\mathcal{P}_j^{(i)}\big\}_{i=1}^\tau\big)\Big)\right),~\forall{ij}\\
c^{(k+1)} &\!\!\!= \mu\left(h-s\right)
\end{dcases}.
\label{eqn:qtSVD-PGA}
\end{equation}
Similarly to the 3DT-DCT case, a $45\degree$ operation was included in a second penalty term $\lambda Q_{45}^{\top} c_{45}^{(k)}$ to $f^{(k+1)}$ in \eqref{eqn:qtSVD-PGA}.

\subsection{Image quality metric}
\noindent The simplest and most and well-known reference quality metrics for images are mean squared error (MSE), or  root MSE (RMSE). These metrics are often too simplistic however, since they do not represent the visual perception of the image very well. In this study, we therefore chose to focus on the structural similarity index (SSIM, 0$\leq$SSIM$\leq$1), since it better captures perceived structural image distortions and noise\cite{Wang2004}. For reconstructed images it is desirable with a large SSIM (relative true) indicating high similarity, and small RMSE indicative of small deviations from true.

\section{Experiments}
\subsection{Simulations}
\noindent Dynamic PET simulations, modeling the General Electric D710 (381 radial and 288 angular bins) were performed using the open source and Matlab-based package dPETSTEP\cite{Haggstrom2016}. This simulation model has been verified to produce highly realistic dynamic PET images. Dynamic simulations of two different phantoms were performed: Dynamic uptake using a spatially stationary brain phantom, and stationary uptake with a spatially dynamic cardiac/lung phantom. The true phantoms will henceforth be referred to as TRUE. Ten noise realizations of each phantom were simulated.

\subsubsection{Cardiac/lung phantom}
\begin{figure}[tb!]
	\centering
		\includegraphics[width=1\columnwidth]{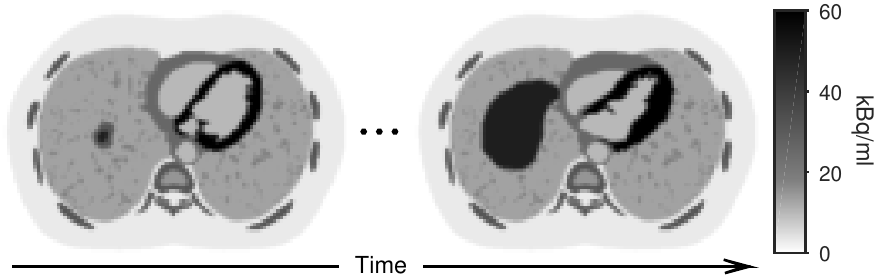}
		\caption{Two out of 150 frames of the cardiac/lung phantom.}
	\label{fig:phantom:heart}
\end{figure}

\noindent The cardiac/lung phantom was constructed from two computed tomography (CT) scans: 1) one stationary anthropomorphic lung phantom scan from the Cancer Imaging Archive\cite{tcia2017}, and 2) a dynamic (1~s cycle over 10 frames of 0.1~s), clinical cardiac CT scan from the OsiriX image library\cite{osirix2017}. A single resampled ($128\times128$) slice of the lung phantom scan was used as a base. The heart masked off and a dynamic slice of the cardiac scan was resampled to the lung phantom voxel size and was inserted in its place. In addition, a dynamic liver (3~s breathing cycle) was also added, resulting in a total of 1 breathing cycle of 30 frames (3~s). The dynamic sequence was repeated to 5 breathing cycles, 150 frames of 0.1~s each, and the activity levels of each tissue were set to realistic $^{13}$N-ammonia values according to scans in the clinic (Memorial Sloan Kettering Cancer Center). The attenuation coefficients for each region ($\mu$-map) was calculated based on the phantom composition. The phantom is seen in \figref{fig:phantom:heart}. The simulation parameters are shown in Table~\ref{tab:simulation}.
\begin{table}[!b]
	\centering
    \caption{Simulation and reconstruction settings.}
		\begin{tabular}{lll}
				\hline\tabularnewline[-0.25cm]
        Item & Cardiac/lung & Brain\tabularnewline[0.05cm]
        \hline\tabularnewline[-0.24cm]
				Average counts/frame  			& 1.5\e{4} (151~kcps) & 1.5\e{5} (1~kcps) \tabularnewline[0.05cm]
				Scatter fraction 						& 0.40 & 0.29 \tabularnewline[0.05cm]
				Random fraction 						& 0.05 & 0.02 \tabularnewline[0.05cm]
				Number of frames						& 150 & 28 \tabularnewline[0.05cm]
				Transaxial FOV							& 400~mm & 256~mm \tabularnewline[0.05cm]
				Matrix size									& $128\times128\times150$ & $256\times256\times28$ \tabularnewline[0.05cm]
				\hline
		\end{tabular}
		\label{tab:simulation}
\end{table}

\subsubsection{Brain phantom}
\begin{figure}[tb!]
	\centering
		\includegraphics[width=1\columnwidth]{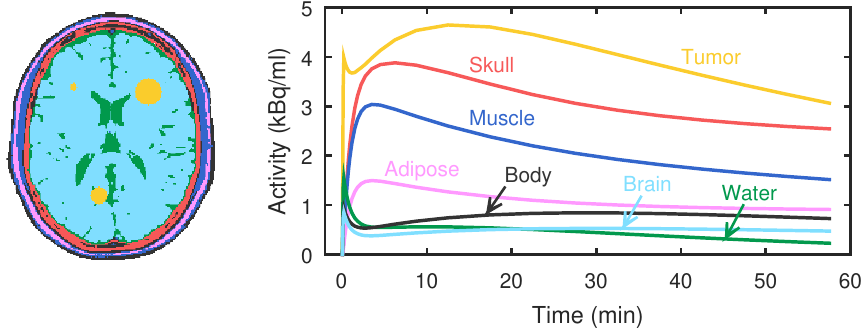}
		\caption{BrainWeb head phantom with three inserted tumor lesions, and associated regional TACs.}
	\label{fig:phantom:head}
\end{figure}

\noindent A $256\times256$ BrainWeb\cite{Collins1998} brain phantom with three inserted lesions was used as input to the dPETSTEP simulation. All regions were assigned realistic kinetic parameters according to the 2-tissue kinetic model for $^{18}$F-fluorothymidine as described in detail previously\cite{Haggstrom2014a,Haggstrom2015,Haggstrom2016}. The $\mu$-map was calculated based on composition. The phantom with the corresponding regional time activity curves (TACs) calculated by dPETSTEP is shown in \figref{fig:phantom:head}. The simulation parameters for this phantom are shown in Table~\ref{tab:simulation}. The total simulation acquisition time was 60~min, divided over 28 frames of $6\times5$~s, $3\times10$~s, $3\times20$~s, $2\times30$~s, $2\times60$~s, $2\times150$~s, $10\times300$~s.

\subsection{Reconstructions}
\noindent The noisy total 3DT projection data, the random+scatter 3DT projection data estimate, and the attenuation factors from the dPETSTEP simulations were used as input to each of the reconstruction algorithms, using settings from Table~\ref{tab:simulation}.
\newline\indent
For 3DT-DCT and 3DT-TNN with patches (Eqs. \eqref{eqn:qDCT-PGA}, \eqref{eqn:qtSVD-PGA}) for the brain phantom, the patch size was set to $8\times8\times4$, and for the and cardiac/lung phantom $8\times8\times150$. The spans were set to $4\times4\times2$, and $4\times4\times150$, respectively.
\newline\indent
All regularized algorithms were iterated until reaching convergence, 
requiring 2200 iterations for the cardiac/lung set, and 600 iterations for the brain phantom. No ordered subsets were used in this work, although they could be incorporated for speed up. For example, Chun \etal\cite{Chun2014} describes a splitting technique for subsets that could be suitable for our patch-based algorithms. Not using subsets, $\beta$ was set to 1. The term $\epsilon$ was set to one hundredth of the median of $f$.
\newline\indent
For each algorithm, optimal weights were found by reconstructing a grid of different $\lambda$, which was stepwise refined up to 10\% change (SSIM unchanged to the third decimal place). The optimal weight was chosen as the one yielding the dynamic image with the highest SSIM relative TRUE.
\newline\indent
The unregularized 2D-OSEM images were reconstructed using 20 iterations and 24 subsets, and post-filtered using a Gaussian filter with full width half maximum (FWHM) that ranged from 0--30~mm (0.5~mm step). The number of iterations and optimal post-filter values were chosen using the images with the highest SSIM. 
\newline\indent
For the cardiac/lung phantom, a cardiac gated image set was also reconstructed. The cardiac cycle was divided into 10 time bins, and each bin was reconstructed with 2D-OSEM. The image set was post-filtered using the same method as for the ungated data.

\subsection{Calculation of LV volume}
\noindent The LV volume for the cardiac/lung phantom images was calculated using Matlab, by region growing from a seed central in the LV. The images were first scaled to a dynamic range of 0--255. LV logical masks were then calculated for a range of different intensity thresholds from 1--30 (growing stops when the intensity difference of a new voxel and the region average exceeds threshold). Holes in the mask were filled. The threshold producing the LV masks (one per frame) with the smallest RMSE relative the true masks was used, and the LV volume was calculated as the sum of mask voxels. We note that LV volumes are normally calculated on volume 3D images. This was nevertheless done here on single image slices as a simple measure of the images clinical usefulness.

\subsection{Parametric image calculation}
\noindent Parametric images were calculated for the brain phantom using dPETSTEP, by weighted nonlinear least-squares (NLS) fitting of the dynamic uptake images of the brain phantom to the 2-tissue model (parameters $K_1$, $k_2$, $k_3$, $k_4$, $V_a$). The true arterial TAC was used as the input function, and initial parameter guesses were set to 0.1. Parametric images were calculated for the conventional 2D-OSEM method, as well as the method found to yield optimal dynamic images. The influx rate was calculated as $K_i=K_1k_3/(k_2+k_3)$\cite{Haggstrom2016}.

\section{Results}
\begin{figure}[tb!]
	\centering
		\includegraphics[width=1\columnwidth]{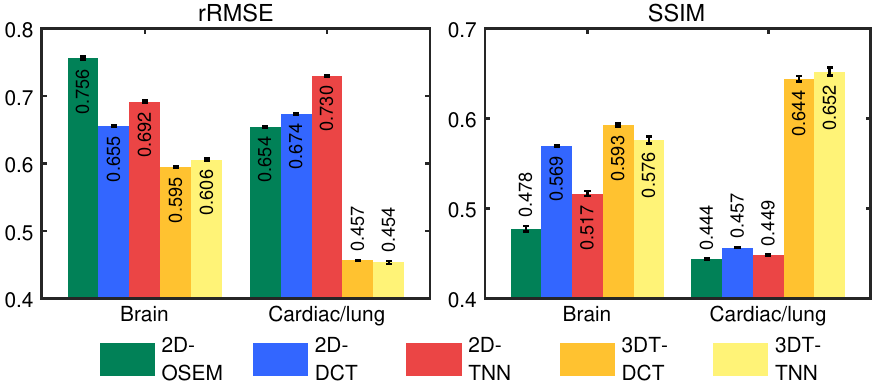}
		\caption{Average rRMSE and SSIM for the different reconstructions. Error bars: 95\% confidence interval (CI).}
	\label{fig:statistics}
\end{figure}

\begin{figure}[bt!]
	\centering
		\includegraphics[width=1\columnwidth]{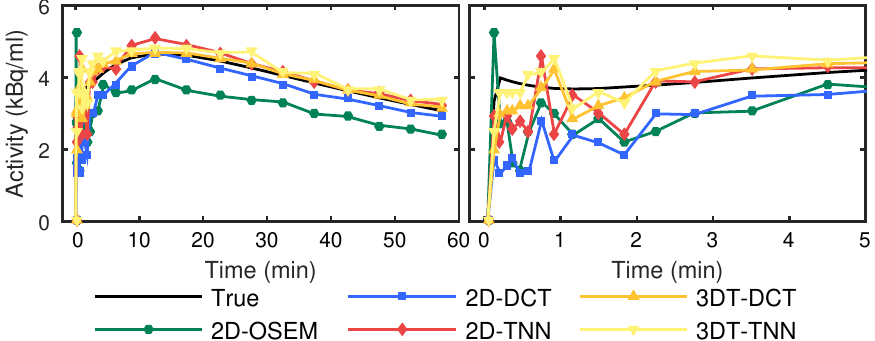}
		\caption{Representative TACs from ROI of the largest tumor of the brain phantom, for the different algorithms. Full data (left), and first 5 min (right).}
	\label{fig:recons:tacs}
\end{figure}

\begin{figure}[bt!]
	\centering
		\includegraphics[width=1\columnwidth]{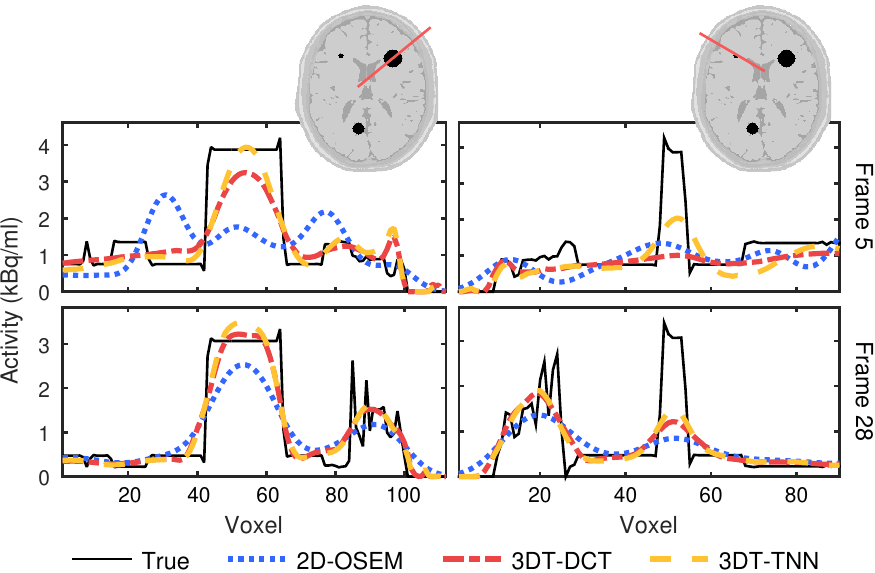}
		\caption{Representative line profiles through two brain phantom tumor lesions, at two different time frames, for conventional 2D and 3DT reconstructions.}
	\label{fig:recons:lineprofiles}
\end{figure}

\begin{figure}[tb!]
	\centering
		\includegraphics[width=1\columnwidth]{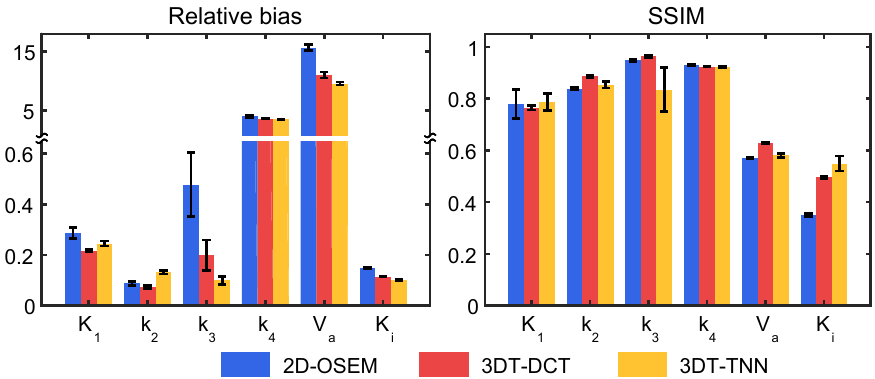}
		\caption{Average relative bias and SSIM for the brain phantom parametric images. Error bars: 95\% CI.}
	\label{fig:recons:pimstats}
\end{figure}

\begin{figure}[t!]
	\centering
		\includegraphics[width=1\columnwidth]{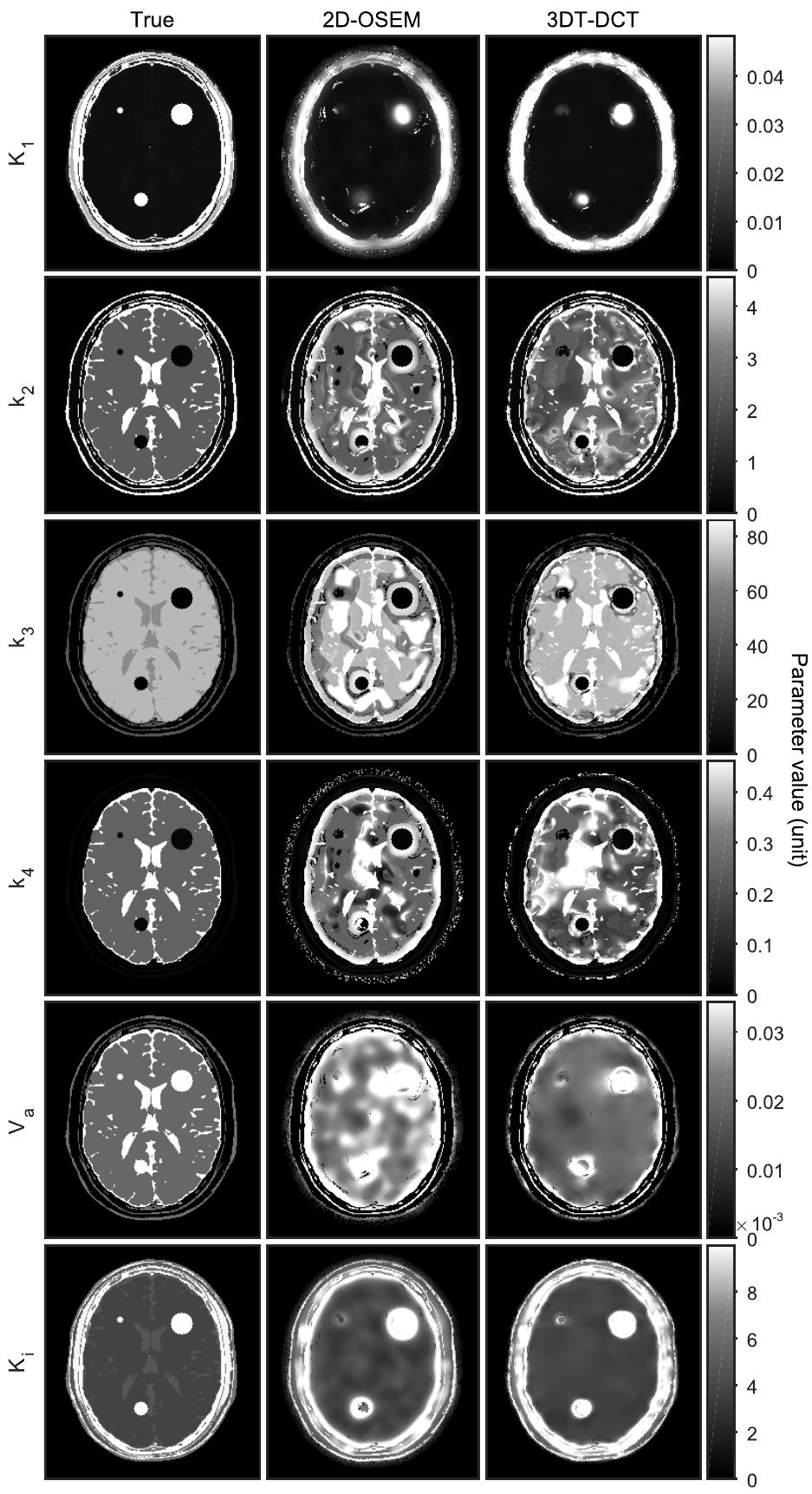}
		\caption{Representative brain phantom parametric images based on conventional and optimal reconstructions. Units $K_1$ and $K_i$ (ml$\cdot$min$\suptext{-1}\cdot$ g$\suptext{-1}$), $k_2$, $k_3$, and $k_4$ (min$\suptext{-1}$), and $V_a$ (ml$\cdot$g$\suptext{-1}$).}
	\label{fig:recons:pim}
\end{figure}

\noindent The optimal post-filter for 2D-OSEM was found to have FWHM 12.5~mm and 27.5~mm for the brain and cardiac/lung phantom, respectively. The FWHM for the gated 2D-OSEM cardiac/lung images was 10.5~mm. Two iterations was optimal for all three sets.
\newline\indent
Values of relative RMSE (rRMSE, relative TRUE average) and SSIM for the different reconstruction methods, averaged across the ten noise realizations, are seen in \figref{fig:statistics}. Bonferroni corrected, one-way anova significance tests show that all rRMSE and SSIM values differ between the different algorithms at the 1\% level, with one exception: SSIM is not significantly different for 2D-OSEM and 2D-TNN for the cardiac/lung phantom.
\begin{figure*}[tb!]
	\centering
		\includegraphics[width=\textwidth]{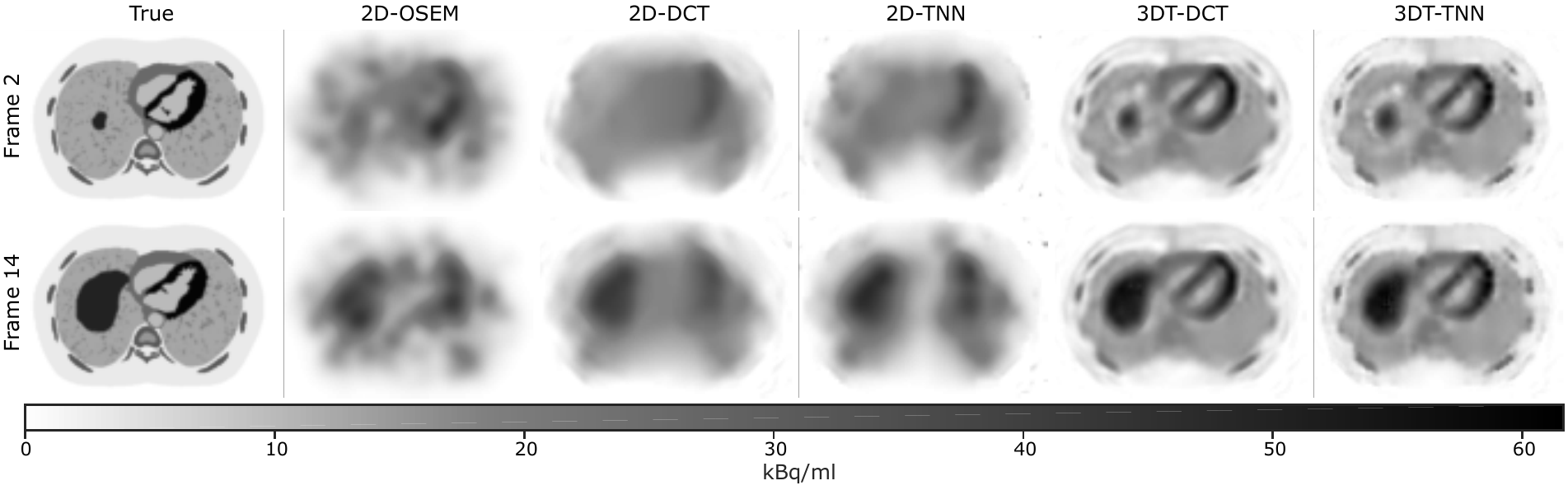}
		\caption{Representative reconstructions of time frame 2 (0.2--0.3~s) and 14 (1.4--1.5~s) of the spatially dynamic cardiac/lung phantom.}
	\label{fig:recons:heart}
\end{figure*}
For the cardiac/lung phantom, the 3DT-TNN w/ patches yielded optimal reconstructions in terms of minimal rRMSE of 45.4$\pm$0.4\%, and maximal SSIM of 0.652$\pm$0.007. 2D-OSEM had corresponding values of 65.4$\pm$0.1\% (rRMSE) and 0.4439$\pm$8\e{-4} (SSIM), respectively. Errors are standard errors (SE).
For the brain phantom, minimal rRMSE of 59.5$\pm$0.3\% was found for 3DT-DCT, and maximal SSIM of 0.593$\pm$0.003. The conventional 2D-OSEM had an rRMSE of 75.6$\pm$0.4\% and an SSIM of 0.478$\pm$0.005.
\newline\indent
\figref{fig:recons:heart} shows representative reconstructions of the cardiac/lung phantom. 
\begin{figure}[tb!]
	\centering
		\includegraphics[width=0.74\columnwidth]{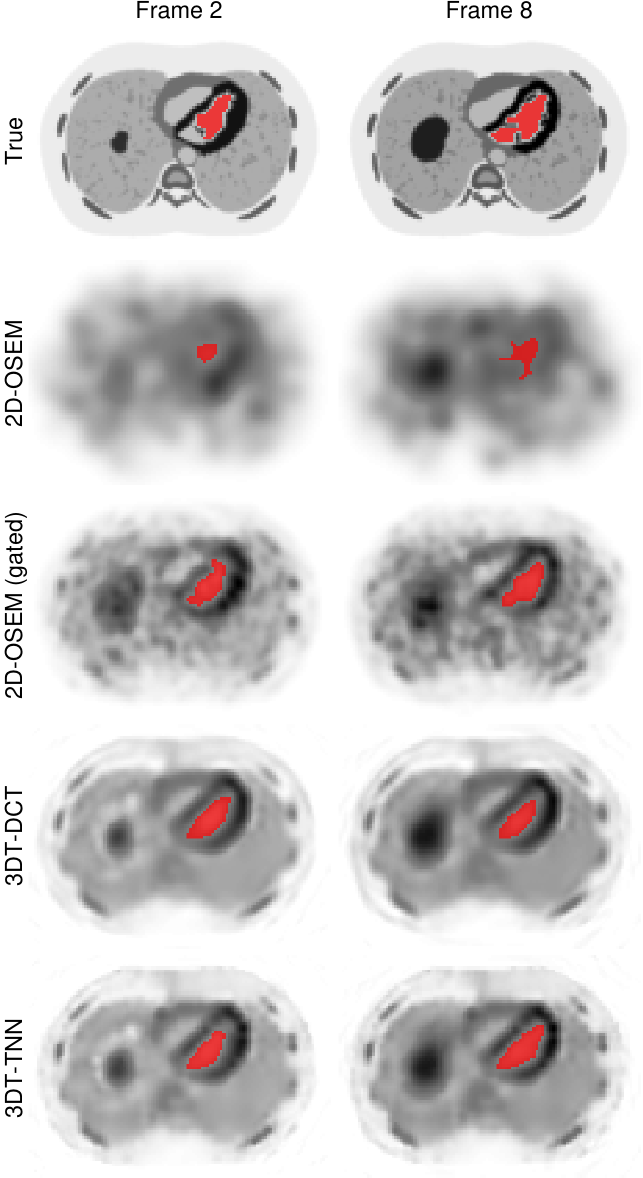}
		\caption{Representative optimal LV masks (red), obtained from region growing.}
	\label{fig:lvmask}
\end{figure}
\begin{figure}[bt!]
	\centering
		\includegraphics[width=1\columnwidth]{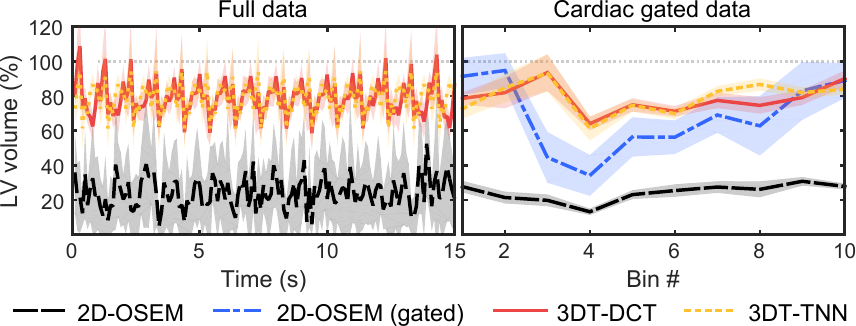}
		\caption{Average region grown LV volume relative true volume. On right, full 2D-OSEM and 3DT-DCT data averaged to bins. Error: 95\% CI.}
	\label{fig:lvvol}
\end{figure}
Representative LV masks are seen in \figref{fig:lvmask}, and the extracted LV volumes from the conventional and optimal image sets are found in \figref{fig:lvvol}. Average (across ten noise realizations) LV volumes of all frame/bin averages are 24.3$\pm$0.6, 68$\pm$1, 79$\pm$2, and 79$\pm$2\% of the true volume, for 2D-OSEM, gated 2D-OSEM, 3DT-DCT, and 3DT-TNN respectively. The average number of misclassified mask voxels were 149.4$\pm$0.5, 81$\pm$2, 74$\pm$2, and 74$\pm$2 for the same four image sets.
As expected, the 3DT-DCT and 3DT-TNN outperforms the conventional 2D-OSEM methods, with superior recovery of the true LV volume by on average 54 and 55 percentage points, respectively, and by 10 and 11 percentage points compared to gated 2D-OSEM.
\begin{figure*}[htb!]
	\centering
		\includegraphics[width=\textwidth]{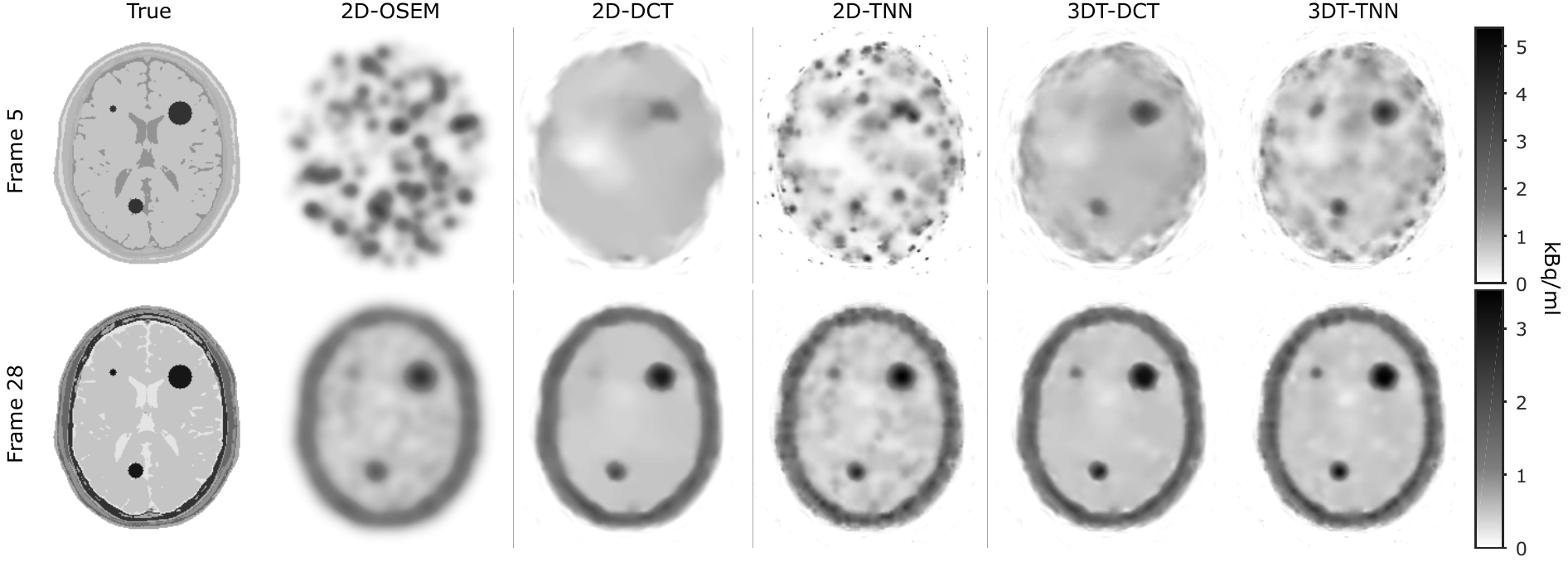}
		\caption{Representative reconstructions of time frame 5 (20--25~s) and frame 28 (55--60~min) of the dynamic uptake brain phantom.}
	\label{fig:recons:brainweb}
\end{figure*}
\newline\indent
Representative reconstructions of the brain phantom are seen in \figref{fig:recons:brainweb}, and TACs from a 5 pixel radius circular ROI central in the largest tumor are found in \figref{fig:recons:tacs}. Line profiles through the largest and smallest lesions are found in \figref{fig:recons:lineprofiles}. 3DT-DCT and 3DT-TNN produce sharper images and are better at recovering the activity of small objects. For example, in frame 28 at 55--60~min (\figref{fig:recons:lineprofiles}, bottom row), 3DT-DCT recovers about 99\% and 39\% of the uptake value for the large and small lesion respectively, compared to the conventional 2D-OSEM at 75\% and 27\%. Frame 5 (20--25~s) has an 75\% and 26\% recovery for 3DT-DCT for the large and small lesion respectively, compared to 41\% and 33\% for 2D-OSEM.
Statistics of brain phantom parametric images calculated from the conventional 2D-OSEM and 3DT reconstructions are found in \figref{fig:recons:pimstats}, and representative images are seen in \figref{fig:recons:pim}. Average bias and SSIM are calculated across the ten noise realizations. For all parameters $K_1$, $k_2$, $k_3$, $k_4$, $V_a$, and $K_i$, relative to 2D-OSEM, 3DT-DCT achieved significantly smaller bias by 7$\pm$1, 1.4$\pm$0.5, 28$\pm$7, 36$\pm$10, 469$\pm$36, 3.2$\pm$0.2 percentage points, respectively. Furthermore, 3DT-DCT achieved larger SSIM for $k_2$, $V_a$, and $K_i$ by 0.047$\pm$0.003, 0.056$\pm$0.002, and 0.145$\pm$0.004 points, whereas it had a smaller SSIM for $k_4$ by 0.006$\pm$0.001 points. SSIM differences in $K_1$ and $k_3$ were not significant. 

\section{Discussion}
\noindent The true dynamic cardiac/lung phantom was created to represent a realistic scan both in movement and activity level. The dynamic brain phantom represented a wide (realistic) variety of contrasts and activity levels (noise) throughout the frames to make the evaluations of the different reconstruction algorithms more comprehensive.
\newline\indent
We have shown that our approaches, incorporating both spatial and temporal penalties, produce higher quality images compared to conventional reconstruction methods. This comes at a cost however, in terms of longer reconstruction times. To be clinically feasible, algorithm speed up is necessary.
\newline\indent
The most straight forward way of speeding up the algorithm is to re-derive it using a list-mode reconstruction framework. This would remove the pointless projections, those that are not paired with any data, which are a feature of high dimensional data. Indeed, the list-mode form of the PG algorithm (or PAPA) reconstructions used here are quite straight forward to derive. However, the dPETSTEP simulation package is not currently compatible with list-mode due to its use of the Matlab Radon transform for forward- and back-projections (the rows and columns of the system matrix are not accessible).
\newline\indent
In this study we have managed to capture both cardiac and breathing motion. Although our phantoms did not include external motion (such as dynamic translation/rotation of the entire phantom) or irregular internal motion, our results are promising for these types of motion as well. Our motion capture method is not dependent on respiratory or cardiac gating, but relies on the assumption of smooth dynamics (physically realistic). The capture of both regular and irregular motion is thus likely feasible. Although regular motion is not a requirement, DCT is especially suitable for regular motion due to its relation to the Fourier transform. 
\newline\indent
The LV volume recovered by region growing on the cardiac/lung images produced significantly better results for our 3DT methods compared to 2D-OSEM. For these types of measurements, it is common practice to use cardiac gated OSEM, and our 3DT DCT and TNN methods outperformed this approach as well. In addition, looking at \figref{fig:lvmask}, although the motion of the cardiac region is successfully captured for the gated images, the breathing motion is not (here seen as the liver movement) and is averaged over many positions causing severe blurring. This is a problem with conventional gating: gating is based on respiratory or cardiac motion, and simultaneous capture is problematic. Gating into both respiratory and cardiac bins, due to even minor irregularity, would result in so many bins that the counting statistics would be extremely low, making reconstruction of reliable images virtually impossible (unless acquisition time is greatly increased - practically not feasible). 
\newline\indent
As seen in \figref{fig:recons:brainweb} and especially \figref{fig:recons:lineprofiles}, the 3DT DCT and TNN algorithms are much better at recovering small objects, compared to conventional post-filtered methods. The line profiles also show that 3DT methods have narrower peaks. The regularized algorithms presented here do not impose smoothing in the same way as ordinary post-filtering does, but denoise images by removing noise components in the image signal. Hence, the loss of resolution is much less for these methods, which is one of their advantages.
\newline\indent
We found that our 3DT-DCT and TNN reconstruction methods followed by NLS compartment model fitting yielded more accurate and precise parametric images for most parameters, compared to the same approach with 2D-OSEM images. Other studies have shown that especially $k_3$ and $K_i$ hold valuable clinical (prognostic) information\cite{Schiepers2007,Spence2009}, and these most important parameters were indeed more accurate and precise using our reconstruction methods.
\newline\indent
At very low count statistics (such as the early frames of the brain simulation at 4--5~kcnts), the 2D results are quite noisy and or over smoothed (\figref{fig:recons:brainweb}). The 3DT-DCT and 3DT-TNN methods yield better results, and at frame rates as low as 1/10~s (15~kcnts) as for the cardiac/lung simulation, images are still rather high quality, especially compared to the 2D methods (\figref{fig:recons:heart}).
\newline\indent
Due to blockiness and hatching artifacts in DCT and TNN, we added an additional image rotated by 45$^{\circ}$ to the penalty (sections \ref{sec:nd-dct}, \ref{sec:nd-tnn}). 
Although this is a heuristic approach, the idea is convincing because the decomposition results in orthogonal rows and columns, which leads to horizontal and vertical enforcement of the penalty. Intuitively this is similar to why isotropic TV semi-norm is generally preferred over the anisotropic one. However, because DCT and TNN is a global penalty (at least for each patch), a larger number of rotationally evenly spaced images may reduce hatching artifacts further, at the expense of more computation. Additional rotations also adds redundancy and over-completeness.
\newline\indent
In this study the temporal patch size included all 150 frames for the cardiac/lung, and 4 frames for the brain phantom. We also reconstructed other patch sizes (results not presented), but yielded inferior results compared to the sizes presented here.
\newline\indent
Commonly, post-filtered 2D (or 3D) reconstruction methods use filters that only incorporate spatial information, \ie frame-by-frame 2D post-filtering. The regular and gated 2D-OSEM results presented here feature just that; 2D Gaussian post-filters. For completeness, we also analyzed the 2D-OSEM data set with optimal 3DT Gaussian post-filtering (results not presented), \ie a Gaussian filter in the $xy$-, and time domains. The recovered LV volume was better than ungated 2D-OSEM, but performed more poorly than gated 2D-OSEM.
\newline\indent
When using minimum rRMSE (relative TRUE) as a metric instead of maximum SSIM, the optimal penalty weights for the different reconstruction algorithms were found to be slightly lower (less strongly penalized). Visually however, the authors found the optimal reconstruction being that with maximal SSIM (as expected).
\section{Conclusion}
\noindent In this study, we investigated dynamic PET image quality, and quality of subsequent extracted image information (LV volume and fitted kinetic parameters) for conventional 2D reconstruction methods as well as novel 3DT methods. We found that our novel 3DT-DCT and 3DT-TNN methods outperform 2D-OSEM in terms of image quality, as well as in calculation of image derived cardiac information and kinetic parameters. In addition, these techniques look promising to help avoid large irregular motion artifacts without need for gating.

\section*{Acknowledgments}
\noindent The authors have no conflicts of interest to report. 
This research was partially supported by the MSK Cancer Center Support Grant/Core Grant P30 CA008748. 

\ifCLASSOPTIONcaptionsoff
  \newpage
\fi



%

%
%
\bibliographystyle{IEEEtran}
\bibliography{References_My_Publications-4D_PETrecon} 

%

%




\end{document}